\documentclass[11pt,a4paper]{article}
\usepackage[hyperref]{acl2020}
\usepackage{times}
\usepackage{latexsym}

\usepackage{graphicx}
\usepackage{todonotes}
\usepackage{subcaption}
\usepackage{makecell}
\usepackage{multirow}

\usepackage{float}
\restylefloat{table}

\usepackage{microtype}

\aclfinalcopy 

\title{Rapid Domain Adaptation for Machine Translation with Monolingual Data}

\author{Mahdis Mahdieh, Mia X. Chen, Yuan Cao, Orhan Firat \\
Google Research \\
\texttt{\{mahdis, miachen, yuancao, orhanf\}@google.com}}

\date{}

\begin{document}
\maketitle
\begin{abstract}
One challenge of machine translation is how to quickly adapt to unseen domains in face of surging events like COVID-19, in which case timely and accurate translation of in-domain information into multiple languages is critical but little parallel data is available yet. In this paper, we propose an approach that enables rapid domain adaptation from the perspective of unsupervised translation. Our proposed approach only requires in-domain monolingual data and can be quickly applied to a preexisting translation system trained on general domain, reaching significant gains on in-domain translation quality with little or no drop on general-domain. We also propose an effective procedure of simultaneous adaptation for multiple domains and languages. To the best of our knowledge, this is the first attempt that aims to address unsupervised multilingual domain adaptation.

\end{abstract}

\section{Introduction}
COVID-19 is an unexpected world-wide major event that hit almost all aspects of human life. Facing such an unprecedented pandemic, how to timely and accurately communicate and share latest authoritative information and medical knowledge across the world in multiple languages is critical to the well-being of the human society. This naturally raises a question of how an existing translation system, usually trained on data from general domains, can rapidly adapt to emerging domains like COVID-19, before any parallel training data is yet available.

Domain adaptation is one of the traditional research topics for machine translation for which a lot of proposals have been made~\citep{chu-wang-2018-survey}. Nevertheless most of them are not suitable for the purpose of rapid adaptation to emerging events. A large body of the existing adaptation approaches are supervised, requiring time-consuming data collection procedure, and while there has been some recent progress made in unsupervised domain adaptation (for example \citep{jin2020simple,dou-etal-2019-unsupervised,dou2020dynamic,hu-etal-2019-domain}), they are not designed specifically to fulfil the requirement of rapidity in domain adaptation, often involving costly algorithmic steps like lexicon induction, pseudo-sample selection, or building models from scratch etc.

In this paper, we propose a novel approach for rapid domain adaptation for NMT, with the goal of enabling the development and deployment of a domain-adapted model as quickly a possible. For this purpose, we keep the following principles in mind when designing the procedure:\\

\noindent\textbf{Simplicity:} The procedure should be as simple as possible, requiring only in-domain monolingual data and avoiding excessive auxiliary algorithmic steps as much as possible.\\

\noindent\textbf{Scalability}: The procedure should be easy to scale up for multiple languages and multiple domains simultaneously.\\
    
\noindent\textbf{Quality:} The adapted model should not sacrifice its quality on general domains for the improvement on new domains.\\

Our approach casts domain adaptation as an unsupervised translation problem, and organically integrates unsupervised NMT techniques with a pre-existing model trained on general domain. Specifically, we engage MASS~\citep{song19d}, an effective unsupervised MT procedure, for the purpose of inducing translations from in-domain monolingual data. It is mingled with supervised general-domain training to form a composite objective in a continual learning setup.

We demonstrate the efficacy of our approach on multiple adaptation tasks including COVID-19~\citep{anastasopoulos2020tico19}, OPUS medical~\citep{TIEDEMANN12.463} as well as an in-house sports/travel adaptation challenge. What is more, we show that this procedure can be effectively extended to multiple languages and domains simultaneously, and to the best of our knowledge, this is the first attempt of unsupervised domain adaptation for multilingual MT.

\section{Background}
\subsection{Unsupervised machine translation} \label{background:unsup}
One of the most intriguing research topics in MT is how to enable translation without the presence of parallel data, for which the collection process is costly. Throughout the history of MT research, many approaches for unsupervised MT have been proposed, but it is not until recent years that significant progress has been made on this topic \citep{artetxe2018unsupervised,lample2018unsupervised,lample-etal-2018-phrase,NIPS2019_8928,artetxe-etal-2019-effective,song19d,liu2020multilingual,Zhu2020Incorporating}, together with the rapid advancement in neural translation models. For example, the BLEU score on WMT14 English-French improved from 15~\citep{artetxe2018unsupervised} to 38~\citep{liu2020multilingual} within just two years.

The approach we propose in this paper, to be detailed in Sec~\ref{sec:approach}, engages unsupervised MT methods for the purpose of domain adaptation. The specific technique we focus on is named MASS~\citep{song19d}, for which we give a brief account as follows. In a nutshell, MASS is an encoder-decoder version of the popular BERT~\citep{devlin-etal-2019-bert} pre-training procedure, in which blocks of the encoder inputs are masked, and are forced to be predicted on the decoder side with only the remaining context available. This procedure is done for monolingual data from both source and target languages, which forces the representation learned for both languages through this denoising auto-encoding process to live in the same space. As a result, even with monolingual inputs, at the end of the MASS training procedure the model's translation ability already starts to emerge. To further boost the translation quality, it is a common practice to continue the training process with online back-translation, which translates target inputs back into source side to form pseudo parallel data to guide model training.

Overall, the algorithm of MASS is simple and elegant while demonstrating superior performance almost comparable to supervised approaches. It naturally fits the encoder-decoder framework and can be easily extended for rapid continual domain adaptation scenario. We therefore adopt this approach as the backbone of our proposed method.

\subsection{Domain adaptation for Machine Translation}
When directly applying an existing NMT system to translation tasks for emerging events like COVID-19, the results often contain numerous errors as the model was never trained on data from this novel domain. The challenging part of this adaptation scenario is that at the beginning of such events, no in-domain parallel corpus is available yet but the NMT system is required to respond properly in time. Therefore an unsupervised and rapid adaptation procedure needs to be in place to fulfil such requirements.

Although domain adaptation has been a traditional research area of MT, most of the existing approaches assume the availability of parallel in-domain data~\citep{freitag2016fast,wang-etal-2017-sentence,zhang-etal-2019-curriculum,thompson-etal-2019-overcoming,saunders-etal-2019-domain,Li2020MetaMTAM}. While there are also approaches that require only monolingual data~\citep{farajian-etal-2017-multi,hu-etal-2019-domain,dou-etal-2019-unsupervised,jin2020simple},, their adaptation procedures are often heavy-weight (for example training data selection, or retrain model from scratch) and not suitable for the purpose of rapid adaptation. What is more, existing approaches usually only consider adaptation towards a single domain for a single language pair. How to rapidly adapt to multiple domains across multiple language pairs remains an under-explored topic.

To address the aforementioned problems, we develop a light-weight, unsupervised continual adaptation procedure that effectively handles multiple domains and languages simultaneously. We now detail our methodology in the following section.

\section{Proposed Approach}
\subsection{Training Procedure Configuration}\label{sec:approach}
\begin{figure}[t]
\includegraphics[width=8cm]{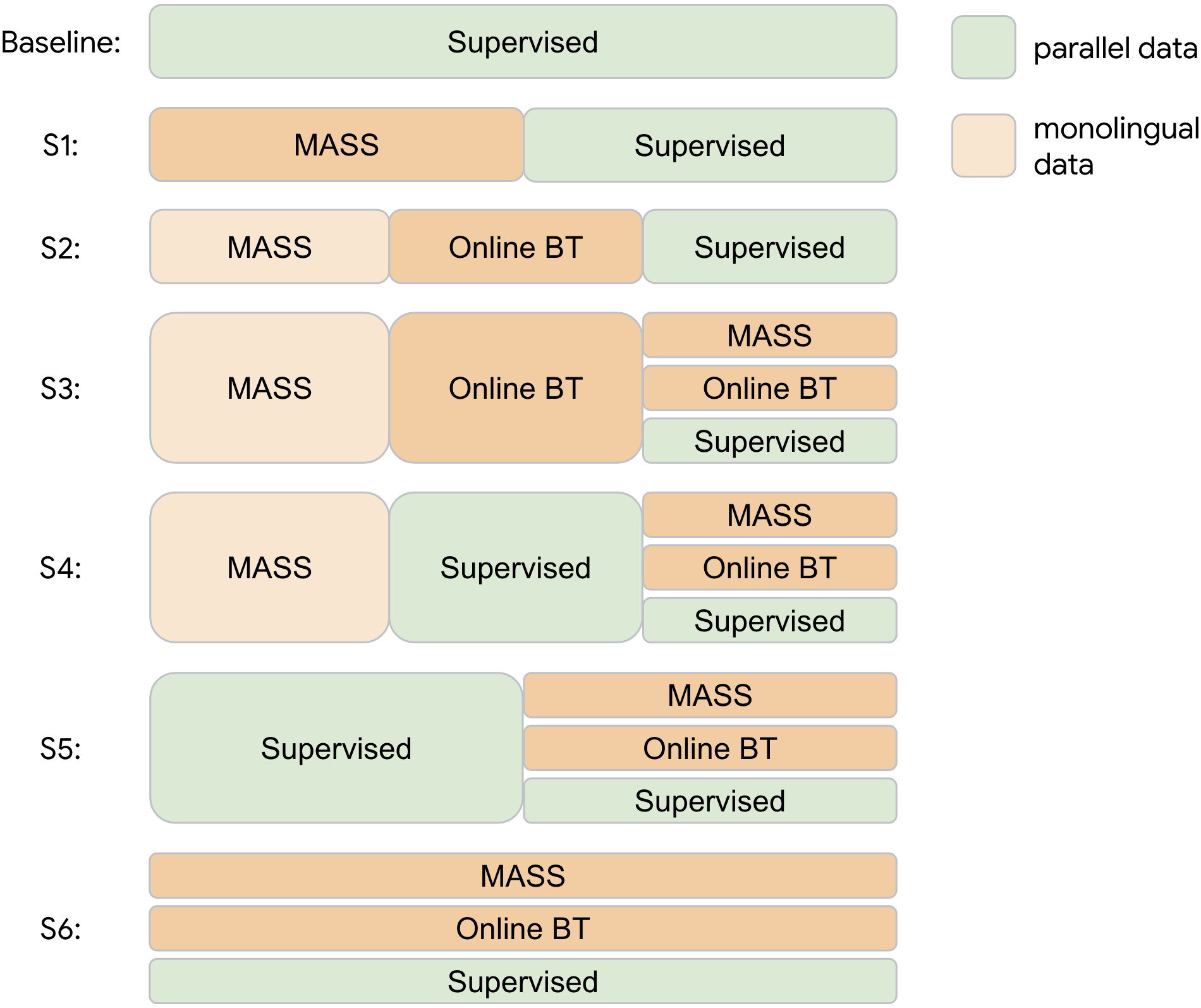}
\caption{Different configurations of the training procedure. Components in light orange, dark orange and green colors are trained with general monolingual data, in-domain monolingual data and general parallel data respectively.}
\label{fig:procedure}
\end{figure}

We treat unsupervised domain adaptation as unsupervised learning of a new language and leverage MASS, introduced in Sec\ref{background:unsup}, as a central building block in our procedure. In order to find out the most suitable way for domain adaptation tasks, we start by investigating different training procedure configurations outlined in Fig~\ref{fig:procedure}. Our training procedures consist of three main components that can be trained sequentially or jointly:





\begin{enumerate}
    \item Supervised training with general parallel data.
    \item MASS pre-training on monolingual data.
    \item Online back-translation using monolingual data.
\end{enumerate}

The monolingual data used for training these components can be either general or in-domain data. Components trained using in-domain data are represented with dark orange color in Fig~\ref{fig:procedure}.

In this paper, we focus on the S4 configuration as it achieves the highest quality improvement on the adapted domain. Also it provides faster domain adaptation compared to other approaches as it only requires in-domain data in the last step of the training process. In section ~\ref{sec:comparison}, we compare these approaches in more details.

S4 consists of three training steps as shown in Fig~\ref{fig:procedure}. The first two steps rely on general parallel and monolingual data, while the third step makes use of in-domain monolingual data. This final step allows us to adapt the model to a new domain rapidly while not suffering from quality loss on the general domain.

\subsection{Multilingual Domain Adaptation}
It has become common for a neural machine translation system to handle multiple languages simultaneously. However, efficiently adapting a multilingual translation model to new domains is still an under-explored topic. We show that our approaches outlined in Sec.~\ref{sec:approach} can be easily extended to multilingual settings.

\subsection{Multi-domain Adaptation}
\label{sec:multi-domain}

Almost all existing work focus on adapting an existing model to one single domain. We explore novel setups where the model is adapted to multiple domains in an unsupervised manner. This provides an insight into the model's ability of retaining previously acquired knowledge while absorbing new information.

With a given general model G, trained using the first two steps of the S4 training procedure, we explore three different setups to adapt G to two new domains A and B:

\begin{enumerate}
    \item G $\rightarrow$ Domain A $\rightarrow$ Domain B
    \item G $\rightarrow$ Domain B $\rightarrow$ Domain A
    \item G $\rightarrow$ $\{$Domain A, Domain B$\}$
\end{enumerate}

Each $\rightarrow$ indicates an adaptation process by jointly training on general parallel data and domain monolingual data based on the third step of the S4 configuration.

\section{Experiments}
\subsection{Setup}

We conduct our experiments on OPUS~\citep{TIEDEMANN12.463} (law and medical domains), COVID-19~\citep{anastasopoulos2020tico19} as well as an in-house dataset in sports/travel domain. For OPUS and COVID-19 experiments, the general-domain parallel and monolingual data comes from WMT, the same corpus as in \citep{siddhant-etal-2020-leveraging}. Detailed dataset statistics can be found in Table \ref{tab:data-stats-parallel} and Table \ref{tab:data-stats-monolingual}. Our in-house datasets are collected from the web. The general-domain parallel data sizes range from 130M to 800M and the sports/travel-domain monolingual data sizes are between 13K and 2M.

We evaluate our approaches with both bilingual and multilingual tasks on each dataset. For OPUS medical and law domains, the bilingual tasks are en$\rightarrow$de, en$\rightarrow$fr, en$\rightarrow$ro and the multilingual task is en$\rightarrow$\{de, fr, ro\}. For COVID-19, they are en$\rightarrow$fr, en$\rightarrow$es, en$\rightarrow$zh and en$\rightarrow$\{fr, es, zh\}. For the in-house sports/travel domain data, we report results on zh$\rightarrow$ja and a 12-language pair (\{en, ja, zh, ko\}$\rightarrow$\{en, ja, zh, ko\}) multilingual model setup.

\begin{table*}[t]
\small
\centering
    \begin{subtable}[h]{0.99\textwidth}
        \centering
        \begin{tabular}{cccccccccc}
        \hline
        \multirow{2}{*}{\textbf{Language Pair}} & \multicolumn{3}{c}{\textbf{General}} & \multicolumn{3}{c}{\textbf{Med Domain}} & \multicolumn{3}{c}{\textbf{Law Domain}} \\
        \cline{2-10}
        & \textbf{Train} &  \textbf{Dev} & \textbf{Test} & \textbf{Train} & \textbf{Dev} & \textbf{Test} & \textbf{Train} & \textbf{Dev} & \textbf{Test} \\
        \hline
        en-de & 4508785 & 3000 & 3003 & 1104752 & 2000 & 2000 & 715372 & 2000 & 2000\\
        en-fr & 40449146 & 3000 & 3003 & 1088568 & 2000 & 2000 & 810167 & 2000 & 2000\\
        en-ro & 610320 & 1999 & 1999 & 990499 & 2000 & 2000 & 451171 & 2000 & 2000\\
        \hline
       \end{tabular}
       \caption{OPUS}
    \end{subtable}
    \hfill
    \begin{subtable}[h]{0.8\textwidth}
        \vspace{1mm}
        \centering
        \begin{tabular}{ccccccc}
        \hline
        \multirow{2}{*}{\textbf{Language Pair}} & \multicolumn{3}{c}{\textbf{General}} & \multicolumn{3}{c}{\textbf{Domain}} \\
        \cline{2-7}
        & \textbf{Train} &  \textbf{Dev} & \textbf{Test} & \textbf{Train} &  \textbf{Dev} & \textbf{Test} \\
        \hline
        en-fr & 40449146 & 3000 & 3003 & 885606 & 971 & 2100\\
        en-es & 15182374 & 3004 & 3000 & 879926 & 971 & 2100\\
        en-zh & 25986436 & 3981 & 2000 & 450507 & 971 & 2100\\
        \hline
       \end{tabular}
       \caption{COVID-19}
    \end{subtable}
    \hfill
    \caption{Statistics of parallel data.}
    \label{tab:data-stats-parallel}
\end{table*}

\begin{table}[h]
\small
\centering
    \begin{subtable}[h]{0.45\textwidth}
        \centering
        \begin{tabular}{ccc}
        \hline
        \multirow{2}{*}{\textbf{Language}}& \multicolumn{2}{c}{\textbf{\# Samples}} \\
        \cline{2-3}
        & \textbf{Med} &  \textbf{Law}\\
        \hline
        en & 1088568 & 810167\\
        fr & 1088568 & 810167\\
        de & 1104752 & 715372\\
        ro & 990499 & 451171\\
        \hline
       \end{tabular}
       \caption{OPUS}
    \end{subtable}
    \hfill
    \begin{subtable}[h]{0.45\textwidth}
        \vspace{1mm}
        \centering
         \begin{tabular}{cc}
        \hline
        \textbf{Language} & \textbf{\# Samples}\\
        \hline
        en & 2315190 \\
        es & 879926 \\
        fr & 885606 \\
        zh & 450507 \\
        \hline
       \end{tabular}
       \caption{COVID-19}
    \end{subtable}
    \hfill
    \caption{Statistics of in-domain monolingual data.}
    \label{tab:data-stats-monolingual}
\end{table}

All the experiments are performed with the Transformer architecture \citep{vaswani2017attention} using the Tensorflow-Lingvo implementation \citep{lingvo}. We use the Transformer Big \citep{chen-EtAl:2018:Long1} model with 375M parameters and
shared source-target SentencePiece model (SPM) \citep{kudo2018sentencepiece} with a vocabulary size of 32k.


\subsection{Results}

\noindent\textbf{Baselines} We compare the results of our proposed unsupervised domain adaptation approach with the corresponding bilingual and multilingual models trained only with general-domain parallel data, without any adaptation. For datasets that have in-domain parallel data available, such as OPUS and COVID-19, we also compare our performance against supervised domain adaptation results, which are produced by experimenting with both continued and simultaneous training using different mixing strategies of in-domain and general parallel data and selecting the best results for each task. In all cases, we report BLEU scores on both general and in-domain test sets.

\vspace{2mm}
\noindent\textbf{Single-domain adaptation} Our bilingual results are shown in Table~\ref{tab:bilingual-onedomain}. Compared with the unadapted baseline models, our unsupervised approach achieves significant quality gain on the in-domain test sets with almost always no quality loss on the general test sets (i.e. learning without forgetting). This improvement is consistent across all three datasets and all languages, with BLEU gains of +13 to +24 on OPUS medical domain, +8 to +15 on OPUS law domain (with the exception of en-fr), +2.3 to +2.8 on COVID-19 and +3.5 on sports/travel domain. Moreover, our method is able to almost match or even surpass the best supervised adaptation performance on a few tasks (e.g., COVID-19 en-fr, en-es, en-zh, OPUS medical en-fr, OPUS law en-ro).

\begin{table}[h]
\small
\centering
    \begin{subtable}[h]{0.45\textwidth}
        \centering
        \begin{tabular}{lccc}
        \hline
        \textbf{Domain} & \textbf{en-de} & \textbf{en-fr} & \textbf{en-ro} \\
        \hline
        Med & \makecell{32.4 (28.2) \\ 45.8 (28.8) \\ 53.5 (28.2)} & \makecell{ 40.4 (38.8)\\ 60.1 (37.2) \\ 63.9 (38.5)} & \makecell{25.8 (26.8) \\ 50.1 (29.3) \\ 59.7 (27.1)} \\
        \hline
        Law & \makecell{43.1 (28.2) \\ 58.2 (29.0) \\ 67.9 (28.1)} & \makecell{60.9 (38.8) \\ 59.5 (38.6) \\ 60.0 (38.7)} & \makecell{35.7 (26.8) \\ 43.9 (29.6) \\ 45.1 (27.4)} \\
        \hline
       \end{tabular}
       \caption{}
    \end{subtable}
    \hfill
    \begin{subtable}[h]{0.45\textwidth}
        \vspace{1mm}
        \centering
        \begin{tabular}{lccc}
        \hline
        \textbf{Domain} & \textbf{en-fr} & \textbf{en-es} & \textbf{en-zh} \\
        \hline
        COVID-19 & \makecell{33.5 (38.8) \\ 35.8 (38.5) \\ 36.3 (38.5)} & \makecell{43.8 (33.7) \\ 46.1 (33.6) \\ 47.9 (33.8)} & \makecell{38.2 (28.3) \\ 41.0 (28.7) \\ 37.8 (28.2)}\\
        \hline
       \end{tabular}
       \caption{}
     \end{subtable}
     \hfill
     \begin{subtable}[h]{0.45\textwidth}
        \vspace{1mm}
        \centering
        \begin{tabular}{lc}
        \hline
        \textbf{Domain} & \textbf{zh-ja}  \\
        \hline
        Sports/Travel & \makecell{21.5 (17.8) \\ 25.0 (17.7)} \\
        \hline
       \end{tabular}
       \caption{}
     \end{subtable}
     \caption{Bilingual, single-domain adaptation results. Table (a), (b), (c) correspond to OPUS medical/law, COVID-19 and in-house sports/travel domain respectively. For Table (a) and (b), each domain contains three rows. The first row represents the baseline model trained with general-domain parallel data without adaptation. The second row is our proposed unsupervised adaptation approach. The third row shows the supervised domain adaptation baseline, serving as upperbound. For sports/travel domain, we do not report supervised adaption results due to lack of in-domain parallel data. In each table cell, the numbers outside and inside parentheses stand for the BLEU scores on the in-domain test set and on the general test set respectively.}
     \label{tab:bilingual-onedomain}
\end{table}

\begin{figure}[h]
\includegraphics[width=8cm]{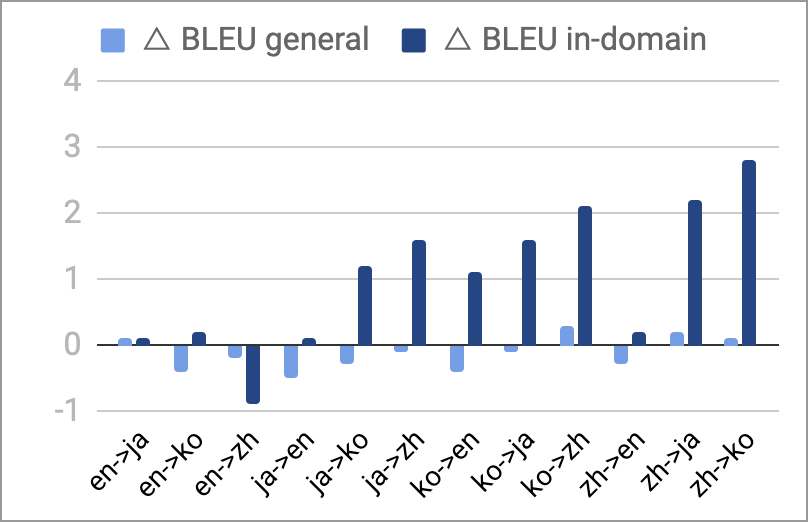}
\caption{BLEU diff on general and sports/travel domain test sets for multilingual single-domain adaptation.}
\label{fig:Multilingual Bleu}
\end{figure}

Table~\ref{tab:multilingual-onedomain} and Figure~\ref{fig:Multilingual Bleu} show our multilingual results. We can see that our approach can be effectively extended to multilingual models. There is large quality improvement across all supported language pairs on the adapted new domains while there is almost no quality regression on the general domains. The improvement ranges from +5 to +9 on OPUS medical domain, +3 to +10 on OPUS law domain, +0.4 to +2.3 on COVID-19 and up to +3 BLEU on sports/travel domain.

\begin{table}[h]
\small
\centering
    \begin{subtable}[h]{0.45\textwidth}
        \centering
        \begin{tabular}{lccc}
        \hline
        \textbf{Domain} & \textbf{en-de} & \textbf{en-fr} & \textbf{en-ro} \\
        \hline
        Med & \makecell{33.7 (28.4) \\ 40.5 (28.6) \\ 56.2 (28.4)} & \makecell{40.0 (37.8) \\ 45.8 (38.1) \\ 63.8 (37.3)} & \makecell{33.6 (30.4) \\ 42.9 (31.0) \\ 60.5 (30.7)} \\
        \hline
        Law & \makecell{43.9 (28.4) \\ 54.6 (28.7) \\ 73.4 (28.2)} & \makecell{57.1 (37.8) \\ 64.3 (38) \\ 39.3 (36.9)} & \makecell{40.9 (30.4) \\ 43.6 (30.8) \\ 46.9 (30.4)} \\
        \hline
       \end{tabular}
       \caption{}
    \end{subtable}
    \hfill
    \begin{subtable}[h]{0.45\textwidth}
        \vspace{1mm}
        \centering
        \begin{tabular}{lccc}
        \hline
        \textbf{Domain} & \textbf{en-fr} & \textbf{en-es} & \textbf{en-zh} \\
        \hline
        COVID-19 & \makecell{33.1 (38.0) \\ 33.5 (38.1) \\ 33.8 (36.6)} & \makecell{44.3 (34.4) \\ 45.8 (34.4) \\ 46.5 (33.6)} & \makecell{36.0 (26.5) \\ 38.3 (28.4) \\ 38.2 (28.1)}\\
        \hline
       \end{tabular}
       \caption{}
     \end{subtable}
     \hfill
     \begin{subtable}[h]{0.49\textwidth}
        \vspace{1mm}
        \centering
        \begin{tabular}{lccc}
        \hline
        \textbf{Domain} & \textbf{en-ja} & \textbf{en-ko} & \textbf{en-zh}  \\
        \hline
        Sports/Travel & \makecell{23.4 (23.6) \\ 23.5 (23.7) } & \makecell{36.9 (40.4) \\ 37.1 (40.0) } & \makecell{30.5 (31.3) \\ 29.6 (31.1) }\\
        \hline
       \end{tabular}
        \begin{tabular}{lccc}
        \textbf{} & \textbf{ja-en} & \textbf{ja-ko} & \textbf{ja-zh}  \\
        \hline
        Sports/Travel & \makecell{22.5 (32.3) \\ 22.6 (31.8) } & \makecell{59.7 (36.1) \\ 60.9 (35.8) } & \makecell{27.9 (26.8) \\ 29.5 (26.7) }\\
        \hline
       \end{tabular}
        \begin{tabular}{lccc}
        \textbf{} & \textbf{ko-en} & \textbf{ko-ja} & \textbf{ko-zh}  \\
        \hline
        Sports/Travel & \makecell{22.4 (28.8) \\ 23.5 (28.4) } & \makecell{45.4 (19.9) \\ 47 (19.8) } & \makecell{29.8 (25.9) \\ 31.9 (26.2) }\\
        \hline
       \end{tabular}
        \begin{tabular}{lccc}
        \textbf{} & \textbf{zh-en} & \textbf{zh-ja} & \textbf{zh-ko}  \\
        \hline
        Sports/Travel & \makecell{21.0 (28.0) \\ 21.2 (27.7) } & \makecell{21.0 (17.7) \\ 23.2 (17.9) } & \makecell{29.4 (34.7) \\ 32.2 (34.8) }\\
        \hline
       \end{tabular}
       \caption{}
     \end{subtable}
     \caption{Multilingual, single-domain adaptation results. Meaning of rows are the same as Table~\ref{tab:bilingual-onedomain}, except that the models are trained and adapted with multilingual setup.}
     \label{tab:multilingual-onedomain}
\end{table}

\vspace{2mm}
\noindent\textbf{Multi-domain adaptation} We demonstrate our multi-domain adaption approaches with a two-domain setup on OPUS medical and law domains. We report the results of the three different setups described in Section \ref{sec:multi-domain} for both bilingual and multilingual scenarios, shown in Table~\ref{tab:bilingual-multidomain} and Table~\ref{tab:multilingual-multidomain} respectively.

Our results suggest that the two-domain simultaneous adaptation approach is able to match the quality of individual single-domain adaptation, with a gap of less than 1.5 BLEU points on both domains and all language pairs for the bilingual models. For the multilingual model, our two-domain adaptation approach matches or outperforms the single-domain adaptation method on the medical domain, while there is a gap of between 0.9 and 4.1 BLEU points on the law domain. Since multi-domain adaptation with a multilingual model requires joint training with both general and in-domain data from all supported language, data mixing/sampling strategy becomes more important in order to achieve balanced quality improvement across multiple domains as well as multiple language pairs.

We further observed that among the three multi-domain adaptation setups, simultaneous adaptation to all domains is the most effective approach. In the sequential setups, there is almost always certain quality regression on the previous domain when the model is being adapted to the second domain.

\begin{table}
\small
\centering
\begin{tabular}{lccc}
\hline \textbf{Pair} &\textbf{Order} & \textbf{Med} & \textbf{Law} \\ \hline
en-de & \makecell{single-domain \\ Med$\rightarrow$ Law\\ Law$\rightarrow$Med \\ \{Med, Law\}} & \makecell{45.8 \\ 38.0 \\ 45.4 \\ 44.8} & \makecell{58.2 \\ 57.6 \\ 46.1 \\ 57.4} \\ \hline
en-fr & \makecell{single-domain \\ Med$\rightarrow$ Law\\ Law$\rightarrow$Med \\ \{Med, Law\}} & \makecell{60.1 \\ 44.0 \\ 59.1 \\ 58.7} & \makecell{59.5 \\ 66.7 \\ 65.6 \\ 65.6} \\ \hline
en-ro & \makecell{single-domain \\ Med$\rightarrow$ Law\\ Law$\rightarrow$Med \\ \{Med, Law\}} & \makecell{50.1 \\ 38.2 \\ 49.8 \\ 48.9} & \makecell{43.9 \\ 43.4 \\ 41.3 \\ 43.5} \\ \hline
\end{tabular}
\caption{\label{tab:bilingual-multidomain} Results of bilingual, two-domain adaptation results. ``Order'' represents the order of the two domains we adapt to during joint training stage, same as the three setups described in Section \ref{sec:multi-domain}. Specifically, $X\rightarrow Y$ indicates adapting to domain $X$ first, then continually adapting to domain $Y$; $\{X,Y\}$ means adapting to domains $X$ and $Y$ simultaneously. ``single-domain'' shows the results of single-domain adaption to each domain as reported in Table~\ref{tab:bilingual-onedomain} (a).}
\end{table}

\begin{table}
\small
\centering
\begin{tabular}{lccc}
\hline \textbf{Pair} &\textbf{Order} & \textbf{Med} & \textbf{Law} \\ \hline
en-de & \makecell{single-domain \\ Med$\rightarrow$ Law\\ Law$\rightarrow$Med \\ \{Med, Law\}} & \makecell{40.5 \\ 37.4 \\ 39.9 \\ 41} & \makecell{54.6 \\ 52.7 \\ 51.5 \\ 50.5} \\ \hline
en-fr & \makecell{single-domain \\ Med$\rightarrow$ Law\\ Law$\rightarrow$Med \\ \{Med, Law\}} & \makecell{45.8 \\ 43.7 \\ 45.3 \\ 46.2} & \makecell{64.3 \\ 62.6 \\ 62.2 \\ 60.6} \\ \hline
en-ro & \makecell{single-domain \\ Med$\rightarrow$ Law\\ Law$\rightarrow$Med \\ \{Med, Law\}} & \makecell{42.9 \\ 40.4 \\ 41.8 \\ 42.7} & \makecell{43.6 \\ 43.1 \\ 42.7 \\ 42.7} \\ \hline
\end{tabular}
\caption{\label{tab:multilingual-multidomain} Results of multilingual, two-domain adaptation results. Meaning of each cell is the same as Table~\ref{tab:bilingual-multidomain}. The single-domain results are from the corresponding multilingual model as in Table~\ref{tab:multilingual-onedomain} (a).}
\end{table}

\subsection{Comparison of Training Procedure Configurations}
\label{sec:comparison}

In this section, we compare the different training procedure configurations described in Section \ref{sec:approach} on the in-house zh$\rightarrow$ja task in sports/travel domain. Table~\ref{tab:bilingual-ablation} shows the best results we were able to obtain for each configuration after experimenting with different data sampling ratios and training parameters. Our main observations are the following:

\begin{itemize}
    \item Comparing with the baseline model, initializing the supervised training stage with a model pretrained using domain monolingual data either with MASS (S1) or both MASS and online back-translation (S2) can result in slight quality improvement (less than 1 BLEU) on the adapted domain.
    
    \item Comparing \{S1, S2\} vs. \{S3, S4, S5, S6\}, joint MASS, online back-translation and supervised training (with both parallel and monolingual data) always seems more effective in boosting the model quality on the adapted domain than purely pipe-lined procedures.
    
    \item It is always helpful to initialize the joint training phase with pretrained models (e.g., S3, S4, S5). Otherwise, it can be hard to find the right sampling ratios among MASS, online back-translation and supervised tasks during a single training process so that the model can improve towards the adapted domain while not having any quality regression on the general domain.
    
    \item Among all the pretraining procedures, it is better to include both MASS and supervised training phases, instead of only supervised training. This way the model would be able to also pick up the language-dependent components inside the architecture during pretraining, which is beneficial for the subsequent joint training phase. Overall, we find that S4 is our most preferable setup. It also offers the advantage of ``rapid'' adaptation, as the MASS and supervised training phases only require general-domain data, thus can be prepared in advance.
    
\end{itemize}

\begin{table}[h]
\small
\centering
\begin{tabular}{cc}
\hline
\textbf{Configuration} & \textbf{test BLEU}  \\
\hline
Baseline & 21.5 (17.8) \\
S1 & 22.3 (17.6) \\
S2 & 22.3 (17.5) \\
S3 & 23.3 (16.4) \\
S4 & 25.0 (17.7) \\
S5 & 23.7 (17.5) \\
S6 & 22.8 (16.4) \\
\hline
\end{tabular}
\caption{Results of the different configurations of the training process on the in-house sports/travel zh$\rightarrow$ja dataset.}
\label{tab:bilingual-ablation}
\end{table}

\section{Related Work}
Domain adaptation is an active topic for MT research~\citep{chu-wang-2018-survey} and has been considered as one of the major challenges for NMT~\citep{koehn-knowles-2017-six}, especially when no or little in-domain parallel data is available.

Perhaps mostly related to our work is~\citep{jin2020simple}, which also relies on denoising autoencoder, iterative back-translation as well as supervision from general domain data for unsupervised domain adaptation. Our work differs from theirs in the following ways: First of all, our work is motivated by rapid adaptation from existing models via continual learning, whereas their work builds in-domain model from scratch, therefore we pay close attention to the prevention of catastrophic forgetting. What is more, we also investigate the problems of simultaneous unsupervised domain adaptation across multiple languages and domains, topics rarely studied before.

While our work is inspired by recent progress made in unsupervised MT, other approaches of using monolingual data for domain adaptation exist. \citep{dou2020dynamic} presents an approach that wisely select examples from general domain data that are representative of target domain and simple enough for back-translation. \citep{dou-etal-2019-unsupervised} propose to use both in- and out-of-domain monolingual data to learn domain-specific features which allow model to specify domain-specific representations of words and sentences. \citep{hu-etal-2019-domain} creates pseudo-parallel training data via lexicon induction from both general-domain parallel data and in-domain monolingual data. \citep{farajian-etal-2017-multi} adapts to any in-domain inputs by selecting a subset of out-of-domain training samples mostly similar to new inputs, then fine-tune the model on this specific subset only for the adaption to the new inputs.

Besides unsupervised domain adaptation, traditionally many approaches have been proposed for supervised domain adaptation. For example model ensembling between in- and out-of-domain models~\citep{freitag2016fast,saunders-etal-2019-domain}, applying regularization that prevents catastrophic forgetting~\citep{thompson-etal-2019-overcoming}, training data selection based on in- and out-of-domain sample similarity~\citep{wang-etal-2017-sentence,zhang-etal-2019-curriculum}, meta-learning for domain-specific model parameters~\citep{Li2020MetaMTAM}.

We also note that our approach is tightly related to techniques for improving NMT quality for low-resource language pairs by making use of monolingual data. For example~\citep{siddhant-etal-2020-leveraging} proposed an approach of improving low-resource translation quality by mingling MASS objective on monolingual data with supervised objectives for high-resource languages during training, and observed significant gains.


\section{Conclusion}
We presented an unsupervised rapid domain adaptation approach for machine translation inspired by unsupervised NMT techniques. Our approach continually adapts an existing model to novel domains using only monolingual data based on a MASS-inspired procedure, which is shown to have significantly boosted quality for unseen domains without quality drop on existing ones. We further demonstrate that this approach is flexible enough to accommodate multiple domains and languages simultaneously with almost equal efficacy. While the problems of domain adaptation, unsupervised and multilingual translation are usually treated as separate research topics, indeed the boundaries between them can be blurred so that a unified procedure can serve all purposes, as our study finds.


\bibliography{acl2020}
\bibliographystyle{acl_natbib}

\end{document}